\documentclass[letterpaper]{article} 
\usepackage{aaai2026}  
\usepackage{times}  
\usepackage{helvet}  
\usepackage{courier}  
\usepackage[hyphens]{url}  
\usepackage{graphicx} 
\usepackage{natbib}  
\usepackage{caption} 
\usepackage{algorithm}
\usepackage{algorithmic}
\usepackage{tcolorbox} 
\usepackage{array}     

\usepackage{newfloat}
\usepackage{listings}
\DeclareCaptionStyle{ruled}{labelfont=normalfont,labelsep=colon,strut=off} 
\floatstyle{ruled}
\newfloat{listing}{tb}{lst}{}
\floatname{listing}{Listing}
\title{The Carbon Footprint Wizard: A Knowledge-Augmented AI Interface for Streamlining Food Carbon Footprint Analysis}

\author {
    Mustafa Kaan Aslan\textsuperscript{\rm 1},
    Reinout Heijungs\textsuperscript{\rm 2},
    Filip Ilievski\textsuperscript{\rm 1}
}
\affiliations {
    \textsuperscript{\rm 1}Department of Computer Science, Vrije Universiteit, Amsterdam, The Netherlands\\
    \textsuperscript{\rm 2}Department of Operations Analytics, Vrije Universiteit, Amsterdam, The Netherlands\\
    m.k.aslan@student.vu.nl, \{r.heijungs,f.ilievski\}@vu.nl
}

\usepackage{bibentry}

\begin{document}

\maketitle

\begin{abstract}
Environmental sustainability, particularly in relation to climate change, is a key concern for consumers, producers, and policymakers. The carbon footprint, based on greenhouse gas emissions, is a standard metric for quantifying the contribution to climate change of activities and is often assessed using life cycle assessment (LCA). However, conducting LCA is complex due to opaque and global supply chains, as well as fragmented data. This paper presents a methodology that combines advances in LCA and publicly available databases with knowledge-augmented AI techniques, including retrieval-augmented generation, to estimate cradle-to-gate carbon footprints of food products. Our methodology is implemented as a chatbot interface that allows users to interactively explore the carbon impact of composite meals and relate the results to familiar activities. A web demonstration showcases our proof-of-concept system with user recipes and follow-up questions, highlighting both the potential and limitations \textemdash such as database uncertainties and AI misinterpretations \textemdash of delivering LCA insights in an accessible format.\end{abstract}

\begin{links}
    \link{Demo}{https://carbonfootprintwizard.labs.vu.nl}
    \link{Code}{https://github.com/mkaanaslan/carbon-footprint-wizard}
\end{links}

\section{Introduction}

As sustainability is high on the agenda for consumers, producers, and policymakers, indicators of sustainable production and consumption are key to present-day research and consulting \cite{krajnc2003indicators}. While sustainability is a multidimensional concept \cite{purvis2019three}, in the context of climate change, environmental considerations take precedence over social and economic factors \cite{yanez2019carbon}. The most popular indicator for quantifying the environmental impact to climate change is the carbon footprint (CF), which aggregates greenhouse gas (GHGs) emissions using global warming potential (GWP) values \cite{CUCEK20129}.\footnote{In this work, we focus on the most recent GWP values from the Intergovernmental Panel on Climate Change (IPCC), with a 100-year time horizon.}

Carbon footprints are typically measured across three scopes \cite{hertwich2018growing}. Scope 1 covers the emissions of GHGs directly from the activity itself, and Scope 2 includes the GHG emissions from the energy production required to run the activity. Scope 3 includes non-energy-related inputs from the whole supply chain, such as materials and processes. Scope 3 is vital for food processes, where it includes issues as diverse as methane emissions from rice cultivation, CO$_2$ emissions from fertilizer production, and methane emissions from cattle breeding for the meat industry. If an analysis includes all three scopes, it takes the form of a life cycle assessment (LCA) \cite{finkbeiner2016introducing}. Most food LCAs are cradle-to-gate (or, with a popular term, farm-to-fork) \cite{schebesta2020european}, stopping short of including the end-of-life phase.

Conducting LCAs is complex due to the long, often global supply chains. To facilitate this, databases such as ecoinvent \cite{wernet2016ecoinvent} and the U.S. Life Cycle Inventory Database \cite{deru2009us} have emerged from early manual studies \cite{pet2003eco,frischknecht1994environmental}.
These databases offer data on either individual unit processes (e.g., operating a tractor), covering materials, transportation, energy, and products, or aggregated data (e.g., the CF of 1 kg of wheat). 
Meanwhile, it is typically more informative for consumers to know the carbon footprint of a compound meal, such as pasta with tomato sauce and parmesan cheese, cooked for 10 minutes. As such data are typically not readily available, producing them from the data for the meal components requires the analyst to fill the gaps, including the materials they are composed of, their quantities, and how the preparation or assembly process works. An illustrative example is an LCA study on pizzas \cite{cortesi_data_2023}.
Moreover, CO$_2$-equivalent values are not always intuitive. For consumers, framing a carbon footprint in relatable terms (e.g., ``like driving 10 km'') is often more meaningful.

To make this data more accessible and user-friendly, we explore using artificial intelligence (AI), particularly large language models (LLMs), augmented with domain knowledge. LLMs, such as the GPT and Claude series, provide natural language interfaces that can process queries in natural language and be used in an interactive, dialogue form. Since LLMs often lack information and may ``hallucinate'' an answer, they can be enhanced with domain-specific documents \cite{lewis2020retrieval} or structured data \cite{edge2024local} through retrieval-augmented generation (RAG) techniques and scripted dialogues.

Recent studies have proposed integrating LLMs with LCA \cite{CORNAGO2023107062,PREUSS2024142824,doi:10.1021/acs.est.4c07634}, often at a conceptual level. Applications are beginning to emerge, e.g., in hydrogen production \cite{make6040122}, construction materials \cite{turhan2023life}, and bio-based materials \cite{turhan2024large}. In parallel, procedures for data retrieval in CF calculations \cite{Luo_Liu_Deng_Yuan_Yang_Xiao_Xie_Zhou_Zhou_Liu_2024,wang2024carbon}, fine-tuning LLMs with CF data \cite{su17031321} have been proposed, estimating food's water footprint \cite{joshi2024knowledge}, or emission factor recommendation \cite{balaji2025emission}. Still, we note a lack of methodologies that effectively model the CF of full meals, reconcile data from multiple overlapping databases, or provide analogs to intuitive processes.

In this paper, we investigate \textit{how the results of LCA studies and existing LCA databases can improve the performance of LLMs in modeling the CF of cradle-to-gate food processes}. 
We develop a methodology that utilizes three freely accessible databases with granular carbon footprint information on food products and other items \cite{bonsai,agribalyse,bigclimate}. Furthermore, we develop a chatbot interface that enables users to explore the carbon footprint of meals interactively and contextualize emission values with familiar activities (e.g., sending an email).

Our solution is implemented as a public demo, 
supported by open-source code.
While still a proof of concept, our method and its implementation demonstrate how AI can help translate complex LCA data into user-friendly, actionable insights, acknowledging that uncertainties remain in both the underlying data and AI interpretation.

\section{Methodology}



\begin{figure*}[h]
    \centering
    \includegraphics[width=1\textwidth]{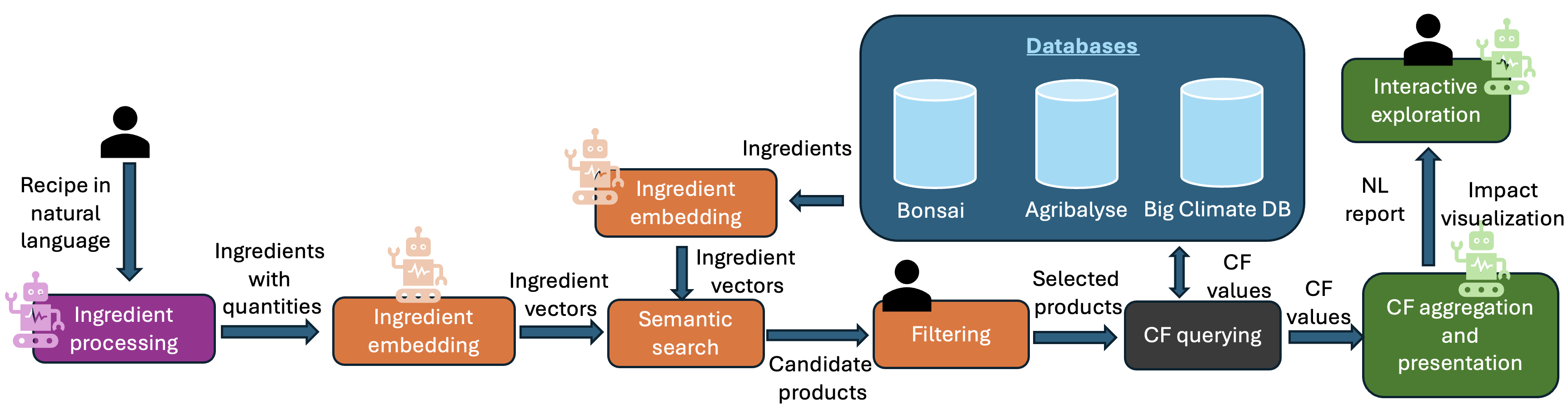}
    \caption{Overview of our four-step methodology. Each step is represented by one color: ingredient processing in purple, product matching in orange, CF querying in black, and iterative exploration in green. The blue container shows the three databases. Arrows indicate data flow. The robot icons indicate LLMs. The person icons indicate input from humans.}
    \label{fig:method}
\end{figure*}

Our proposed methodology implements an integrated pipeline to generate environmental impact assessments from free-form recipe inputs, designed to overcome key challenges in food sustainability analysis. As illustrated in Figure \ref{fig:method}, the methodology consists of four stages: 
\begin{enumerate}
    \item \textbf{Ingredient processing}, which uses an LLM to extract structured ingredient data from textual descriptions;
    \item \textbf{Product matching}, which uses vector embeddings to match ingredients with database entries for products despite naming variations, followed by user verification;
    \item \textbf{CF querying}, which retrieves CF values from the corresponding databases;
    \item \textbf{Interactive exploration}, which synthesizes data from multiple LCA databases into comprehensive impact reports and visualizations, and enables the user to continue to chat with the large language model.
\end{enumerate}

Existing LCA approaches often fail to serve non-expert users as they lack readily available data for compound meals and require specialized skills to assemble carbon footprints from individual ingredients. Our system addresses these limitations by leveraging AI techniques to interpret natural language queries and match them with LCA data, while maintaining human oversight at critical decision points. This hybrid approach combines AI and humans, making complex LCA data accessible to laypeople while ensuring high quality through human validation of crucial mapping decisions. We describe our procedure in this section, whereas the exact prompts for the LLMs are provided in the Appendix.

\subsection{LCA Databases}

\begin{table}[t]
\centering
\small
\begin{tabular}{l c c c}
\hline
\textbf{Database} & \textbf{BONSAI} & \textbf{Agribalyse} & \textbf{Big Climate DB} \\
\hline
\textbf{\# Products} & 411 & 2,616 & 540 \\
\textbf{\# Categories} & - & 11 & 13 \\
\textbf{Geo Focus} & Global & France & Europe (5) \\
\hline
\end{tabular}
\caption{Number of food products and categories, and geographical focus of the three data sources in our framework.}
\label{tab:data_sources_numeric}
\end{table}

Our system leverages environmental impact data from three key LCA databases: BONSAI~\cite{bonsai}, Agribalyse~\cite{agribalyse}, and Big Climate Database~\cite{bigclimate}. These databases were chosen primarily for their open-source availability, which ensures transparency and accessibility, while also providing robust, high-quality data critical for evaluating the carbon footprints of food products. Each brings distinct advantages that enhance the accuracy, granularity, and contextual relevance of our environmental impact assessments.

\paragraph{BONSAI.} BONSAI is a global open-source database that provides multiregional input-output data, allowing detailed assessments of food product emissions in diverse geographical regions. Its hierarchical structure links products, activities, and recipes, offering a fine-grained view of supply chain emissions, from raw material extraction to market delivery. Inclusion of recipe-specific data further allows for precise tracking of inputs and CO\textsubscript{2} equivalent outputs in agricultural and manufacturing processes, making it an invaluable tool for complex analysis. Still, its coverage of regions and products remains incomplete, with full development expected by June 2025.

\paragraph{Agribalyse.} Agribalyse is an open-source database tailored to the French food sector, providing comprehensive life cycle inventory (LCI) and impact data for more than 2,600 food products. It excels in covering the entire lifecycle—agriculture, processing, packaging, transportation, retail, and consumption—using a standardized LCA methodology that ensures consistency and reliability. A notable feature is its data quality rating for each entry, which gives confidence in the results and supports informed decision-making. However, its focus on French consumption patterns may limit its applicability elsewhere without adaptation.

\paragraph{Big Climate Database.} It provides open-source, region-specific environmental impact data for 540 food products in five European countries: Denmark, the United Kingdom, France, Spain, and the Netherlands. Its standout feature is the incorporation of indirect impacts of land use change (iLUC), offering a broader perspective on the environmental consequences of food production. Detailed emission breakdowns across the lifecycle stages—agriculture, processing, packaging, transport, and retail—further enhance their utility for targeted analysis. Although its product scope is narrower than Agribalyse's, its regional precision compensates for this limitation.

\subsection{Ingredient Processing}
\label{ssec:ingredient}

Given a user input, the first step is converting unstructured recipe inputs into a standardized format suitable for database matching and impact calculation. 
The accuracy of this initial extraction step is crucial, as errors or omissions at this stage can spread throughout the entire pipeline, potentially leading to significant inaccuracies in the final impact assessment.
Natural language recipe descriptions present several challenges. Ingredients may be described with varying levels of detail (e.g., ``minced beef'' versus ``beef''), quantities can be specified in various units (cups, tablespoons, pieces), and key information might be embedded within cooking instructions or casual language (e.g., "a handful of spinach"). 

To address these challenges, we opt for an LLM agent with a carefully engineered prompt that guides the extraction process.
Specifically, we utilize an LLM through a function-calling interface that enforces structured output. The LLM is prompted with explicit instructions to identify ingredient names and quantities, with the requirement to convert all measurements to grams using standard culinary conversions (e.g., 1 tablespoon of oil $\approx$ 15g). The model returns a JSON object containing an array of ingredient objects, each with name and quantity fields. The resulting ingredient names and their corresponding amounts in grams are then stored and serve as input for the subsequent semantic search phase.

\subsection{Product Matching}
\label{ssec:search}

Establishing accurate correspondences between user-provided ingredients and database entries presents a significant challenge due to variations in naming conventions across different LCA databases. Direct string matching fails to capture critical semantic relationships - for instance, "minced beef" and "ground beef" refer to the same product but would be considered distinct under exact matching. Semantic search techniques address this limitation by matching ingredients based on their meaning rather than their exact textual representation. Semantic search leverages vector space representations that capture the contextual relationships between different food items, enabling the system to identify relevant products even when their names differ significantly from the user's input.

We use an embedding model to generate vector embeddings for both ingredient names and database entries. These embeddings capture semantic relationships between food items based on their contextual usage patterns in large text corpora. We maintain a pre-computed index for each product name in the database (BONSAI, Agribalyse, and Big Climate Database), storing normalized embeddings for all product names. The ingredient names are transformed into vectors when processing a user query using the same model and normalization process. The system then performs efficient similarity searches using these indices to retrieve the top 3 most similar products from each database, employing cosine similarity as the distance metric. This approach ensures that semantically related products are matched appropriately while maintaining computational efficiency through optimized nearest neighbor search algorithms. 

Although semantic search provides a robust method for identifying potential product matches, the inherent complexity of food products and variations in naming conventions require human oversight in the final selection process. Our method includes an interactive product selection interface, where users are presented with up to three candidate products from each database for each ingredient in their recipe. For each ingredient, the system displays the ingredient name, quantity, and a list of semantically similar products identified from BONSAI, Agribalyse, and the Big Climate Database, using checkboxes. For instance, both "ground beef" from BONSAI and "minced beef" from the Big Climate Database might be displayed when searching for a "beef" ingredient. Users can select multiple products for each ingredient to account for cases where different databases use different names for the same product or when a combination of products better represents the intended ingredient. If the user cannot find a suitable match, no results are provided for that ingredient. The selected products are assigned to their corresponding database identifiers and passed to the next stage for impact calculation. If a user selects a product for which data is not available for the target country, we calculate an average impact value using the available data from other countries. This human-in-the-loop approach strikes a balance between automation and accuracy, ensuring that the final environmental impact assessment is based on correctly matched products while maintaining an efficient and user-friendly workflow.

\subsection{Carbon Footprint Querying}
\label{ssec:query}

From BONSAI, we extract both production-level emissions data and market-based emissions that account for regional production shares. Specifically, when querying BONSAI for a selected product, the system retrieves direct production emissions, where available, followed by a comprehensive breakdown of market composition that shows percentage contributions from different geographic origins. Each market share entry includes the region name, percentage value, and associated emissions data, preserving the multi-tiered data model essential for understanding global supply chains. From Agribalyse, we gather total impact values expressed in kg CO\textsubscript{2}-equivalent, along with their data quality ratings and detailed lifecycle stage breakdowns that partition impacts into six distinct phases: agriculture, processing, packaging, transportation, retail, and consumption. The Agribalyse query process preserves both the absolute values and the proportional contribution of each lifecycle stage. From the Big Climate Database, we retrieve region-specific total impact values and their detailed breakdowns across production phases, with a particular focus on their unique components, such as indirect land-use change impacts, which are not captured in other databases. The Big Climate Database query also includes identifying regional variations when data for the target country is unavailable.

The system structures this information into a standardized results text as input for the LLM's analysis. It creates a section for each ingredient, starting with its name and quantity, followed by results from each available database. When BONSAI data are available, we first list any direct production emissions, followed by a detailed breakdown of market shares and their associated emissions for different regions. For Agribalyse matches, we include the total impact value, data quality rating, and percentage contributions from each stage of the lifecycle. Similarly, for Big Climate Database matches, we present the total impact and phase-specific contributions, including unique indirect land-use change impacts. All impact values are consistently expressed in kilograms of CO\textsubscript{2} equivalent per specified ingredient quantity, enabling straightforward comparison and aggregation during the analysis step.

\subsection{Interactive Exploration}
\label{ssec:chatbot}

The last phase of the system uses an LLM to integrate structured environmental impact data from various databases into a cohesive evaluation and presentation of a recipe's CF. This process is systematically executed through a series of well-defined steps that ensure accuracy, consistency, and transparency in the analysis. The LLM interprets structured data extracted from multiple environmental databases using a carefully crafted prompt that provides explicit instructions for parsing data formats unique to each source. For example, the prompt enables the LLM to calculate market-based total impact values from BONSAI, lifecycle-based aggregated values from Agribalyse, and region-specific values from the Big Climate Database, using the extracted data. The prompt also specifies protocols for reconciling data from multiple sources, such as calculating minimum and maximum impact ranges and derived average values. In addition, it includes guidelines for estimating cooking-related impacts when relevant, ensuring that LLM can unify heterogeneous data into a consistent analytical output.

The recipe's environmental impact computation follows predefined rules in the LLM's prompt. Per ingredient, the LLM determines a range of impact values (minimum to maximum) when multiple values are available, or adopts a single value when data are sourced from a single database. When the recipe implies cooking processes (e.g., baking, boiling), the LLM estimates supplementary impacts based on standard energy consumption metrics associated with the cooking method, duration, and temperature. The total environmental impact is then calculated by aggregating the impacts of all ingredients and any associated cooking processes, resulting in both a range and an average value. To enhance transparency, the system flags deviations from the user-specified target country, ensuring that the analysis remains contextually relevant.

The LLM generates a structured output comprising a natural language summary and accompanying visualization data. The summary ranks the recipe's ingredients by carbon footprint and presents the total impact range and average. Moreover, it contextualizes these findings with relatable comparisons to everyday activities (e.g., number of emails that cause a similar footprint), calibrated against reference values manually extracted from the book ``How Bad Are Bananas?'' \cite{berners2020bad}. The visualization data supports the creation of graphical representations, including a horizontal bar chart and a pie chart. These visuals delineate the proportional contributions of each ingredient and cooking process to the overall carbon footprint, ensuring that the results are presented in a clear and standardized manner that supports user comprehension.

In its initial response, the system presents only the most critical information: ingredient impacts ordered by magnitude, cooking considerations, total recipe impact, and relatable comparisons. However, the LLM maintains access to all detailed data, including regional market shares, lifecycle stage breakdowns, and granular emissions data across different stages of production and distribution. This selective presentation approach prevents information overload while allowing for iterative, in-depth exploration through follow-up questions. Users can ask for specific aspects, such as market share distributions, regional variations, or lifecycle patterns, and the LLM draws upon the complete dataset to provide detailed responses. This creates an interactive experience where users can progressively explore the complex aspects of their recipe's environmental impact through natural dialogue.

\begin{figure*}[!ht]
    \centering
    \includegraphics[width=0.84\textwidth]{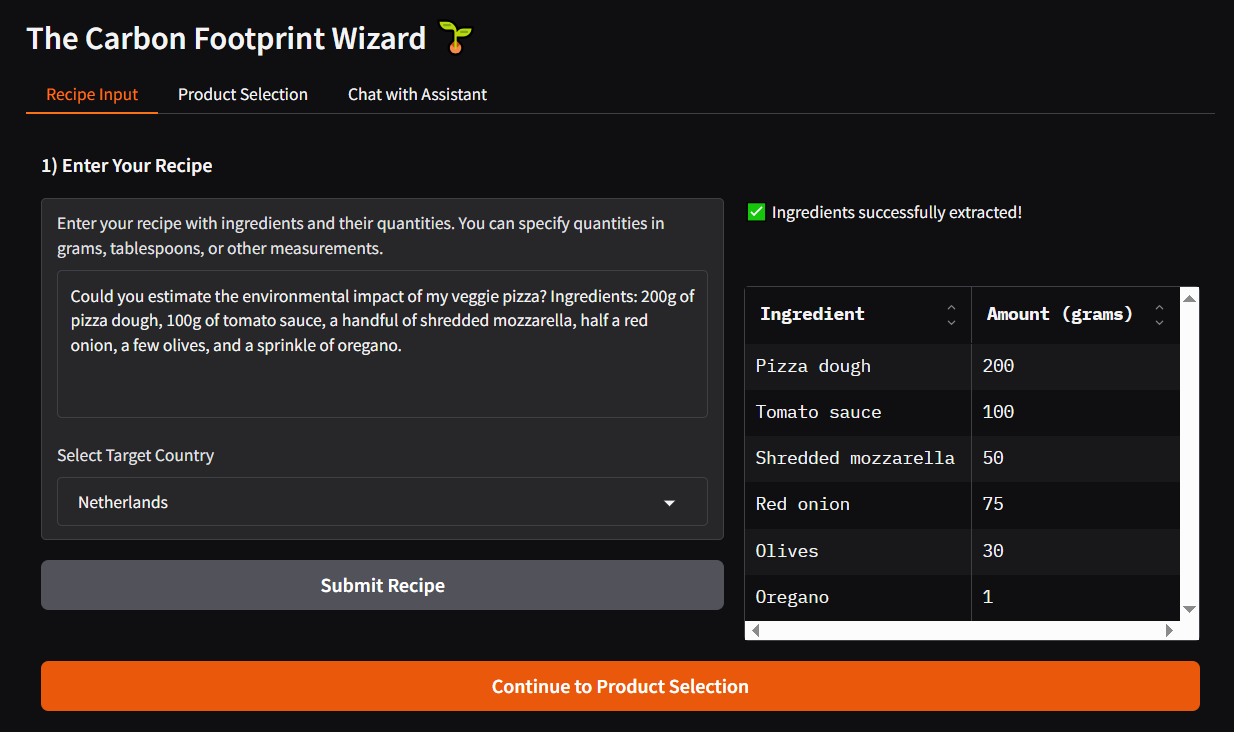}
    \caption{Recipe input interface with ingredient processing from natural language (shown on the left) to standardized quantities (organized as a two-column table on the right).}
    \label{fig:extraction}
\end{figure*}

\subsection{Implementation Details}

The methodological pipeline is implemented as our \textbf{Carbon Footprint Wizard} tool, which utilizes a combination of AI tools to enable efficient processing, matching, and analysis of recipe data. Ingredient processing leverages Open AI's GPT-4o-mini \cite{hurst2024gpt}, configured with a temperature parameter of zero for deterministic outputs, to extract structured ingredient lists from natural language inputs via a function-calling interface that enforces JSON formatting. For semantic search, the all-MiniLM-L6-v2 Sentence Transformer model generates vector embeddings for ingredient names and database entries, capturing semantic relationships.
\footnote{\url{https://huggingface.co/sentence-transformers/all-MiniLM-L6-v2}. Accessed: 2025-04-17.} 
The resulting embeddings are then indexed and queried efficiently using an FAISS index \cite{douze2024faiss} to retrieve the most similar products based on cosine similarity. Database querying and result synthesis are facilitated by custom functions that interface with BONSAI, Agribalyse, and Big Climate Database. The chatbot interface uses GPT-4o-mini to process structured impact data and generate natural language responses and visualization data with zero temperature setting, integrated into a Gradio-based \cite{abid2019gradio} web interface for user interaction. For real-world equivalence scaling, we extracted eight entries from daily life activities in ``How bad are bananas?'' \cite{berners2020bad}, such as sending an email (0.004 kg CO\textsubscript{2}-eq), driving 1 mile in a Fiat 500 (0.35 kg CO\textsubscript{2}-eq), and hand washing dishes (8.0 kg CO\textsubscript{2}-eq), specifically chosen to help users better understand and contextualize the CF of the recipe through relatable comparisons. Visualization of the CF, presented as a horizontal bar chart and a pie chart, is implemented using the Matplotlib library \cite{hunter2007matplotlib}, which allows for intuitive and standardized graphical representations of ingredient impacts.

\section{Case Study Analysis}

We show the effectiveness of our methodology and the corresponding Carbon Footprint Wizard tool through a step-by-step analysis using a vegetarian pizza recipe. 
We do not perform a quantitative evaluation because there are no ground-truth values for the CF of the meals.
We outline the step-by-step workflow with the outputs of our system, highlighting its ability to provide detailed environmental impact assessments from natural language recipe descriptions.

Figure \ref{fig:extraction} illustrates the initial \textbf{ingredient processing} stage of our interface, where the user submits a recipe description in natural language: \textit{"Could you estimate the environmental impact of my veggie pizza? Ingredients: 200g of pizza dough, 100g of tomato sauce, a handful of shredded mozzarella, half a red onion, a few olives, and a sprinkle of oregano."} The user selects the Netherlands as the target country for the analysis. The Carbon Footprint Wizard's LLM successfully extracts and standardizes the ingredients, converting qualitative descriptions into quantified amounts: \textit{pizza dough (200g), tomato sauce (100g), shredded mozzarella (75g), red onion (70g), olives (30g), and oregano (5g)}. This demonstrates the system's ability to interpret ambiguous quantities (e.g., "a handful," "half," "a few," "a sprinkle") and convert them into standardized weight measurements in grams.

After ingredient extraction, the system performs \textbf{ingredient matching}. For this purpose, the tool's semantic search retrieves and presents multiple potential product matches for each ingredient to the user through the product selection interface, as shown in Figure \ref{fig:selection} in the Appendix. For example, \textit{"pizza dough"} returns seven distinct matches including "bread," "mixes and doughs," "pastry," "pizza base, cooked," "pizza base, raw," "pizza dough", and "pizza with tomato and cheese, ready meals". Similarly, for \textit{"shredded mozzarella"}, options include "cheese from skimmed cow milk", "cheese, semihard, mozzarella, 30\% fdm", "mozzarella cheese, from cow's milk", and "processed cheese, in slices". The interface reveals the complexity of ingredient classification across databases: \textit{"red onion (70g)"} yields numerous options including "garlic", "onion, cooked", "onion, dried", "onion, raw", "onions, dry", and "red onion", with only the exact match "red onion" selected by the user as the most adequate one. For \textit{"olives"}, the system provides specialized options like "olives, black, without stones, in brine" alongside generic "olives", with both selected to capture comprehensive data. The interface also demonstrates the breadth of alternative product classes that could be semantically related but conceptually distinct, such as "tapenade" as an option for \textit{"olives"} and "cream sauce with herbs" as an option for \textit{"oregano"}. This deliberate selection process addresses the challenge of varying terminology across different databases while allowing users to match ingredients based on their specific recipe knowledge. The asterisk notation indicates products without country-specific data, informing users that estimates from other countries will be used in such cases. This human-in-the-loop approach ensures accuracy in product matching while maintaining an intuitive user experience. Using the selection from the user, the \textbf{CF querying} step is performed in the background.

\begin{figure*}[!ht]
    \centering
    \includegraphics[width=0.83\textwidth, trim={0 0 0 1cm},clip]{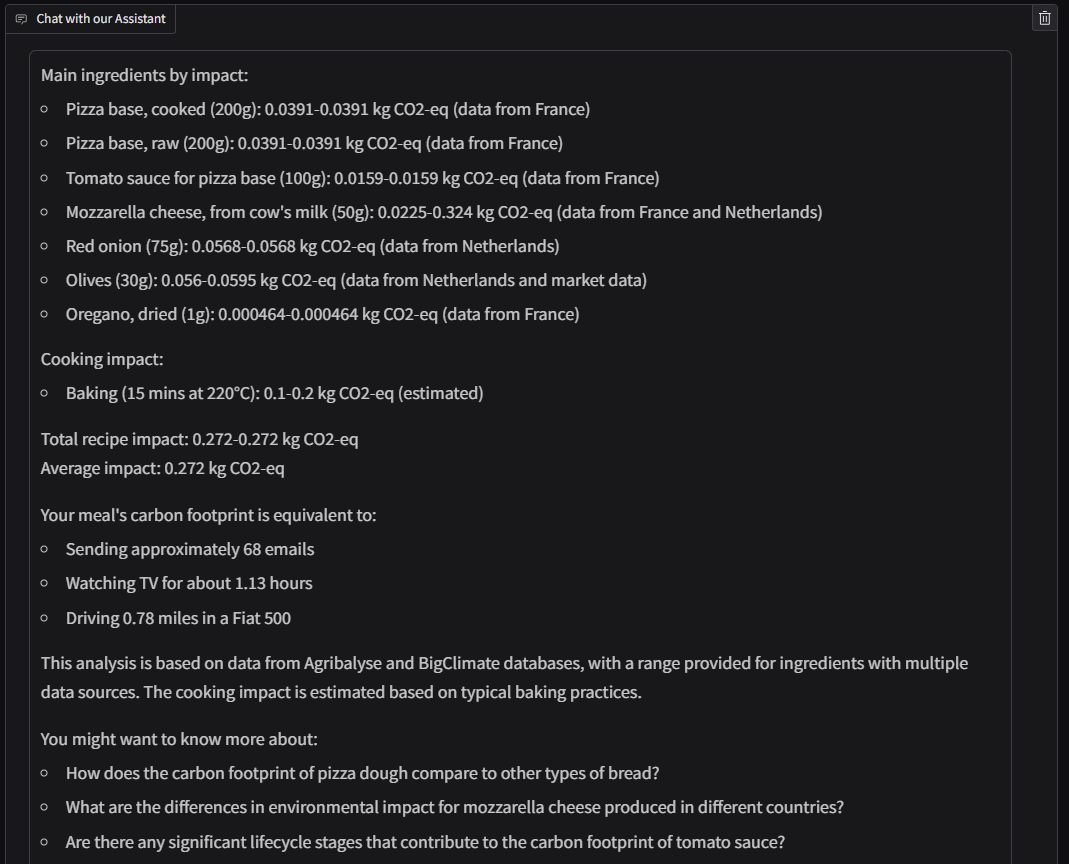}
    \caption{Initial analysis showing ranked ingredients, impact range, and real-world equivalents as a natural language summary.}
    \label{fig:initial_result}
\end{figure*}

Following user selection and database querying for CF values, the LLM performs a comprehensive CF analysis and visualization in the \textbf{interactive exploration} phase. The natural language report shown in Figure \ref{fig:initial_result} presents the ingredients ranked by their environmental impact, with mozzarella cheese identified as the highest contributor, followed by pizza base, olives, red onion, tomato sauce, and oregano in descending order. For each ingredient, the system notes when the numbers are estimated based on data from countries other than the user-selected target country (the Netherlands), enhancing transparency in data provenance. While the system should typically estimate cooking impacts even when not explicitly mentioned in the recipe, in this specific case, it incorrectly reports no cooking impact (0 kg CO\textsubscript{2}-eq) for the pizza preparation, thus violating common sense. The analysis provides both a total recipe impact range (0.14-0.84 kg CO\textsubscript{2}-eq) and calculates an average impact value of 0.49 kg CO\textsubscript{2}-eq as the final carbon footprint estimate, which represents only the ingredients without cooking emissions. To contextualize these values, the system translates the carbon footprint into equivalent everyday activities: sending approximately 122 emails, watching TV for about 2 hours, or driving 1.4 miles in a Fiat 500. This comparison helps users conceptualize the environmental impact in relatable terms. Additionally, the system suggests potential follow-up questions to promote further exploration of the results.

The visual representation of the carbon footprint analysis in Figure \ref{fig:visuals} features a horizontal bar chart and a pie chart. These visualizations provide a clear breakdown of each ingredient's contribution to the total environmental impact, allowing users to quickly identify which recipe components have the largest carbon footprints. The charts reveal that mozzarella cheese (49\%) and pizza base (26\%) are the dominant contributors to the carbon footprint. At the same time, tomato sauce contributes only 3\% despite making up a substantial portion of the recipe by weight.

Figure \ref{fig:interact} demonstrates the interactive capabilities of the system, showing a follow-up inquiry on the market shares of ingredients. In response, the system provides detailed information on the geographical origins of each ingredient. Although market share data are unavailable for most ingredients, olive data reveal market share from various countries, including Italy, Germany, and Portugal. This interaction showcases the system's ability to provide granular details from the underlying databases not included in the initial summary, demonstrating the depth of information available through the conversational interface. On the flip side, it shows the importance of having high-quality LCA databases that fuel the tool in the background.

\section{Discussion}

The Carbon Footprint Wizard effectively bridges the gap between complex environmental impact databases and user-friendly insights. By combining natural language processing for ingredient processing, user-assisted semantic search for product matching, and a conversational interface for interactive result exploration, the system makes LCA data accessible to non-expert users.
At the same time, our analysis reveals three key challenges with missing data, inconsistencies between the databases, and violations of common sense by the LLM. 

First, the Carbon Footprint Wizard reveals challenges regarding \textbf{data availability}. The three databases used (BONSAI, Agribalyse, and Big Climate Database) exhibit notable gaps in country-specific product coverage, which necessitates relying on data from other regions in many cases. This geographical mismatch is evident in the user interface, where asterisks denote products that lack data for the selected country. In our pizza example, several ingredients showed data from France instead of the requested data from the Netherlands, potentially reducing the accuracy of region-specific assessments. 

Second, \textbf{inconsistencies} between the footprints reported in the databases result in considerable impact ranges rather than precise values, as evidenced by the wide min-max intervals (0.13-0.84 kg CO\textsubscript{2}-eq) presented in the results. This variability is partly due to inherent differences in how each database handles lifecycle phases and system boundaries. Another aspect of data consistency concerns the lack of standardized vocabulary used in the different databases. The semantic matching process, while generally effective, occasionally suggests irrelevant products (such as "tapenade" for olives or "cream sauce with herbs" for oregano), reflecting the inherent challenges of grounding natural language ingredient descriptions into standardized database terminology. While our current methodology employs a human-in-the-loop approach to mitigate this issue (by allowing users to select appropriate matches), it introduces dependency on user expertise and judgment, creating another potential source of variability in the final assessment results.

Finally, the system also demonstrates limitations in \textbf{handling complex analytical tasks using a single LLM}. For instance, the cooking impact estimation occasionally fails to detect implied cooking requirements, as demonstrated by the veggie pizza example, where the system incorrectly indicated "no cooking required" despite pizza preparation typically involving baking. Such commonsense violations can potentially lead to a significant underestimation of the total environmental impact of the recipe, as cooking can contribute substantially to the CF of a meal. Similarly, the generation of follow-up questions sometimes falls short of identifying the most relevant aspects for further exploration, focusing on generic queries rather than insights specific to the analyzed recipe's unique characteristics. These issues stem from the ambitious scope of responsibilities assigned to a relatively small language model (GPT-4o-mini), including ingredient extraction, impact calculation, cooking estimation, visualization, data preparation, scaling real-world equivalence, and follow-up question generation. The model must simultaneously interpret heterogeneous data formats from multiple databases, perform mathematical operations, generate natural language explanations, and create structured visualization data, all within a single inference pass. This overextension of the model's capabilities could be addressed in future iterations by agentic AI, which can decompose these tasks into separate, specialized LLM calls that focus on distinct aspects of the analytical pipeline. This approach may involve employing larger and more capable models for the most complex reasoning tasks.

\section{Conclusions}

This paper introduced an interdisciplinary methodology that bridges state-of-the-art LCA frameworks and databases with the current generation of knowledge-augmented AI. The methodology's four steps leveraged LLMs for seamless user interaction, fueled by LCA databases with contextual knowledge about food products. The methodology was designed to involve users in the ingredient selection step and to allow users to explore the data further after the summarized report was presented. The methodology is implemented in the publicly available Carbon Footprint Wizard web interface. The initial analysis of the tool's functionality showcased the power of connecting LCA insights with LLMs, providing an intuitive entry point to complex information for lay users. At the same time, the analysis revealed challenges with missing data, inconsistencies between the databases, and violations of common sense by the LLM. These challenges point to substantial interdisciplinary obstacles that can be fruitfully explored in the future: the engineering and representation of LCA knowledge, the development of LCA data quality frameworks, and ensuring consistency in LLM behavior. Agentic AI frameworks and advanced entity resolution approaches can be employed to address these limitations. Future work should address two limitations of our approach. First, the lack of quantitative validation of our method necessitates benchmarking against expert LCAs, conducting sensitivity/error analyses, and conducting user studies to demonstrate accuracy, robustness, and usability at scale. Second, our three databases differ in geographical scope, system boundaries, and category definitions, leading to wide impact ranges and limited comparability. Thus, future work should revise the selection of LCA databases to account for their differences in coverage and focus.

\section*{Acknowledgements}



This work has been supported by the seed grant ``Knowledge-based support for sustainable diet policies'' awarded to the authors by the Vrije Universiteit’s Amsterdam Sustainability Institute (ASI).


\bibliography{aaai2026}


\clearpage
\appendix

\section{Prompts Used in the System}
\label{app1}

\subsection{Ingredient Extraction Prompt}

\begin{tcolorbox} 
{\footnotesize
Extract the ingredients and their quantities from the following user message.

If the quantities are provided in units other than grams (such as tablespoons, cups, or pieces), convert them to a reasonable weight in grams based on typical measurements.

You can directly change milliliters to grams.

Provide the ingredients and quantities in the format required by the function 'process\_ingredients'.

\textbf{User Message:} \{user\_message\}
}
\end{tcolorbox}

\subsection{Result Generation Prompt}

\begin{tcolorbox} 
{\footnotesize 
You will receive two inputs: \\
\textbullet~user\_message: Original recipe query \\
\textbullet~results\_text: Impact data from databases \\
\textbf{DATA FORMAT OVERVIEW:} \\
\textbullet~BONSAI: Shows market/production data with country shares. Use market total impact value. \\
\textbullet~Agribalyse: French data with lifecycle stages. Use total impact value. \\
\textbullet~BigClimate: Country-specific data with indirect land use. Use total impact value. \\

\textbf{CALCULATION RULES:} \\
1. Per ingredient: \\
- Multiple databases: Use min-max range and (min+max)/2 average \\
- Single database: Use value for both \\
- Note if data isn't from target country \\
- Use market data from BONSAI when available \\

2. Recipe total: \\
- Range: sum(min) to sum(max) \\
- Average: sum(individual averages) \\
- Cooking analysis: \\
    \textbullet~Determine if the recipe requires any form of cooking for preparation \\
    \textbullet~If cooking is required: \\
        - Estimate cooking impact range (min-max) \\
        - Add cooking range to total recipe range \\
        - Add cooking average to total average \\
        - Include cooking in visualization data \\
    \textbullet~If no cooking required: state "No cooking required (0 kg CO$_2$-eq)" \\

}
\end{tcolorbox}


\begin{tcolorbox} 
{\footnotesize
3. After impacts, calculate equivalent activities: \\
- Use total average (Z kg CO$_2$-eq) with these references: \\
    \textbullet~Sending an email = 0.004 kg CO$_2$-eq \\
    \textbullet~Web search on a laptop = 0.0007 kg CO$_2$-eq \\
    \textbullet~Watching TV 42-inch plasma = 0.24 kg CO$_2$-eq \\
    \textbullet~Driving 1 mile Fiat 500 = 0.35 kg CO$_2$-eq \\
    \textbullet~3-minute shower = 0.09 kg CO$_2$-eq \\
    \textbullet~Charging phone daily = 0.003 kg CO$_2$-eq \\
    \textbullet~Using laptop 1 hour = 0.05 kg CO$_2$-eq \\
    \textbullet~Hand washing dishes = 8.0 kg CO$_2$-eq \\
- Present 2-3 scaled comparisons \\

4. List some follow up questions: \\
- Generate 3-4 most relevant follow-up questions based on: \\
    \textbullet~Market share of ingredients \\
    \textbullet~Notable data variations between countries \\
    \textbullet~Interesting lifecycle patterns \\
    \textbullet~Potential impact reduction opportunities \\

}
\end{tcolorbox}

\begin{tcolorbox}
\footnotesize
\textbf{OUTPUT FORMAT:} \\
- Main ingredients by impact: \\
- [Ingredient] ([amount]g): [X-Y kg CO$_2$-eq] (note if data is from a different country) \\
- Cooking impact: \\
- [Method] ([time] mins at [temp]°C): [X-Y kg CO$_2$-eq] \\
- Total recipe impact: [X-Y kg CO$_2$-eq] \\
- Average impact: [Z kg CO$_2$-eq] \\

Your meal's carbon footprint is equivalent to: \\
- [Scaled comparisons] \\

[Brief paragraph: data sources, range explanation, cooking estimate] \\

You might want to know more about: \\
- [List the follow-up questions] \\

\textbf{CRITICAL:} You must return TWO components in your response as the defined "process\_impact\_results": \\
1. answer\_user: Follow the output format above \\
2. visualization\_data: Must include these exact fields: \\
   - ingredients: Array of ingredient names from the user recipe (including cooking if applicable) \\
   - impacts: Array of corresponding impact values in kg CO$_2$-eq (use average when range exists) \\
   - DO NOT include total recipe impact on visualization! \\

After the initial response, user will continue chatting with you. Do not use function format after initial response, answer directly. \\

Here is user message: \\
\{user\_message\} \\

Here is the information from our sources: \\
\{results\_text\}
\end{tcolorbox}

\section{Example Outputs from Each Step}
\label{app2}

Below are the example outputs from each step of the environmental impact analysis process for a vegetarian pizza recipe.

\subsection{User Recipe Input}

\begin{tcolorbox} 
{\footnotesize
Could you estimate the environmental impact of my veggie pizza? Ingredients: 200g of pizza dough, 100g of tomato sauce, a handful of shredded mozzarella, half a red onion, a few olives, and a sprinkle of oregano.
}
\end{tcolorbox}

\subsection{Ingredient Extraction}

\begin{tcolorbox} 
{\footnotesize
\begin{tabular}{|l|r|}
\hline
\textbf{Ingredient} & \textbf{Amount (grams)} \\
\hline
Pizza dough         & 200 \\
Tomato sauce        & 100 \\
Shredded mozzarella & 75  \\
Red onion           & 70  \\
Olives              & 30  \\
Oregano             & 5   \\
\hline
\end{tabular}
}
\end{tcolorbox}

\subsection{Product Matching}

\begin{tcolorbox} 
{\footnotesize
\begin{tabular}{|l|p{0.6\textwidth}|}
\hline
\textbf{Ingredient} & \textbf{Matching Products} \\
\hline
Pizza dough & Bread, Mixes and doughs, Pastry, Pizza base, cooked *, Pizza base, raw *, Pizza dough, Pizza with meat, tomato and cheese, ready meals, Pizza with tomato and cheese, ready meals, Pizza, cheese and mushrooms * \\
\hline
Tomato sauce & Chili sauce, Cream sauce *, Juice of tomatoes, Soya sauce, Tomato sauce for pizza base *, Tomato sauce, with onions, prepacked *, Tomatoes, fresh, Tomatojuice, canned \\
\hline
Shredded mozzarella & Cheese from skimmed cow milk *, Cheese, semihard, mozzarella, 30 \% fidm., Mozzarella cheese, from cow's milk *, Plant-based cheese, without soybean, prepacked, shredded *, Processed cheese, Processed cheese, in slices *, Tomato, peeled, canned, Vegan cheese, grated or sliced, Whey cheese \\
\hline
Red onion & Garlic, Onion, cooked *, Onion, dried *, Onion, raw, Onions, chopped and deepfried, Onions, dry, Onions, shallots (green), Red onion \\
\hline
Olives & Oil of olives, virgin, Olive oil, Olive oil, extra virgin *, Olives, Olives, black, without stones, in brine, Olives, green, pickled, canned, Pizza, onion anchovy and black olives *, Tapenade (a puree of capers, pitted black olives, anchovy and herbs, with olive oil and lemon juice) * \\
\hline
Oregano & Cream sauce with herbs *, Garlic, Garlic, raw, Ginger, Olive oil, Oregano, dried *, Rosemary, dried *, Soya paste, Sunflower oil \\
\hline
\end{tabular}
}
\end{tcolorbox}

\begin{figure*}[!ht]
    \centering
    \includegraphics[width=1\textwidth]{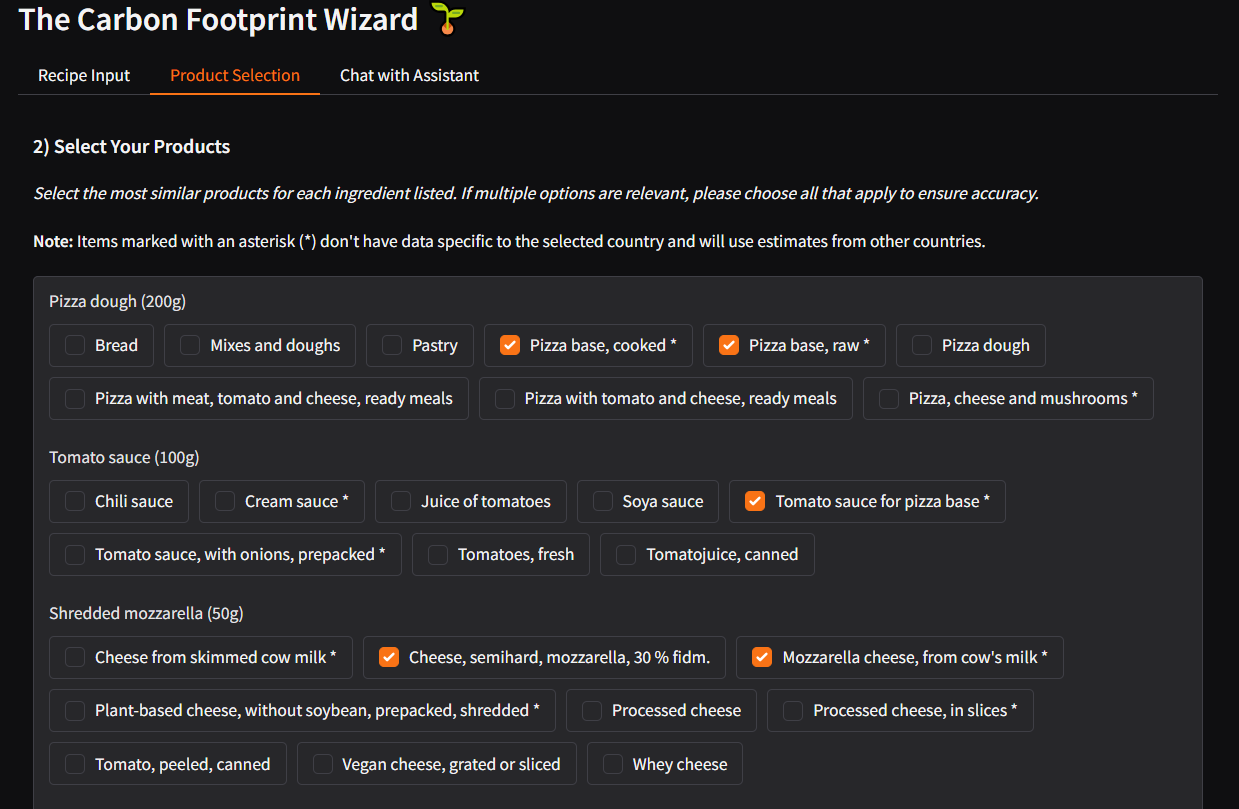}
    \caption{The product selection showing multiple database matches for each ingredient, resulting from the semantic search in the product matching phase.}
    \label{fig:selection}
\end{figure*}

A screenshot of the matching step in our interface is shown in Figure \ref{fig:selection}.

\subsection{Carbon Footprint Data}

\begin{tcolorbox}
{\footnotesize
The results text is a structured compilation of environmental impact data for each selected product from the databases (e.g., AgriBalyse, BigClimateDatabase, BONSAI). It includes sections for each ingredient, with detailed data such as total impact and contributions from various lifecycle stages. Below is an excerpt for the 'pizza dough' ingredient:

\textbf{Results for selected most similar items to 'pizza dough':} \\

\textbf{Agribalyse database results for 'Pizza base, raw' (DATA FROM FRANCE):} \\
- Impact for 200 grams: 0.0391 kg CO$_2$-eq \\
- Data quality rating: 2.3277205962237506 \\
- Agriculture impact for 200 grams: 0.00267 kg CO$_2$-eq, Percentage: 6.8\% \\
- Processing impact for 200 grams: 0.00128 kg CO$_2$-eq, Percentage: 3.3\% \\
- Packaging impact for 200 grams: 0.00121 kg CO$_2$-eq, Percentage: 3.1\% \\
- Transportation impact for 200 grams: 0.000711 kg CO$_2$-eq, Percentage: 1.8\% \\
- Retail impact for 200 grams: 0.000381 kg CO$_2$-eq, Percentage: 1.0\% \\
- Consumption impact for 200 grams: 0.0014 kg CO$_2$-eq, Percentage: 3.6\% \\

\textbf{BigClimateDatabase results for 'Pizza dough' in Netherlands:} \\
- Impact for 200 grams: 0.241 kg CO$_2$-eq \\
- Agriculture impact for 200 grams: 0.174 kg CO$_2$-eq, Percentage: 72.2\% \\
- Indirect Land Use Change impact for 200 grams: 0.011 kg CO$_2$-eq, Percentage: 4.6\% \\
- Food processing impact for 200 grams: 0.00465 kg CO$_2$-eq, Percentage: 1.9\% \\
- Packaging impact for 200 grams: 0.0354 kg CO$_2$-eq, Percentage: 14.7\% \\
- Transport impact for 200 grams: 0.0155 kg CO$_2$-eq, Percentage: 6.4\% \\
- Retail impact for 200 grams: 0.000312 kg CO$_2$-eq, Percentage: 0.1\% \\

Similar data is provided for all ingredients (tomato sauce, shredded mozzarella, red onion, olives, oregano) in the recipe, detailing impacts from relevant databases.
}
\end{tcolorbox}

\begin{figure*}[!ht]
    \centering
    \includegraphics[width=\textwidth, trim={0 0 0 1cm},clip]{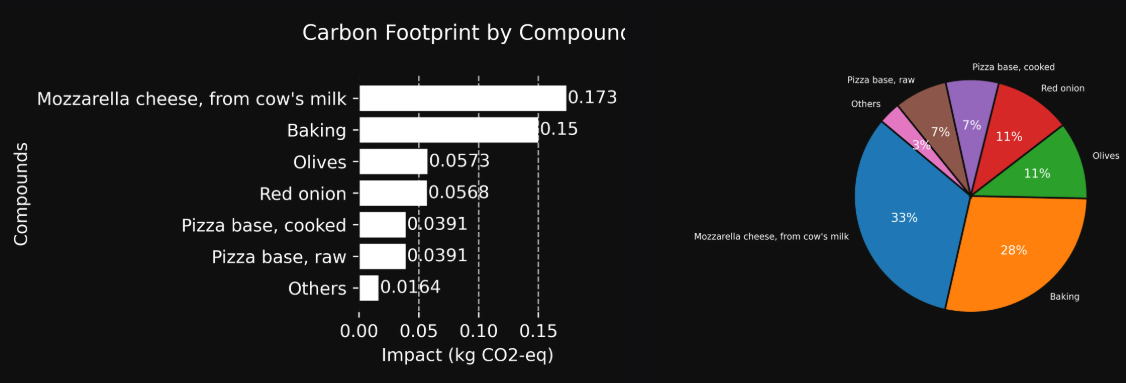}
    \caption{Carbon footprint visualization through bar and pie charts showing ingredient impacts.}
    \label{fig:visuals}
\end{figure*}

A visual summary of the analysis results provided by our interface is shown in Figure \ref{fig:visuals}.

\subsection{Natural Language Report}

\begin{figure*}[!ht]
    \centering
    \includegraphics[width=1\textwidth]{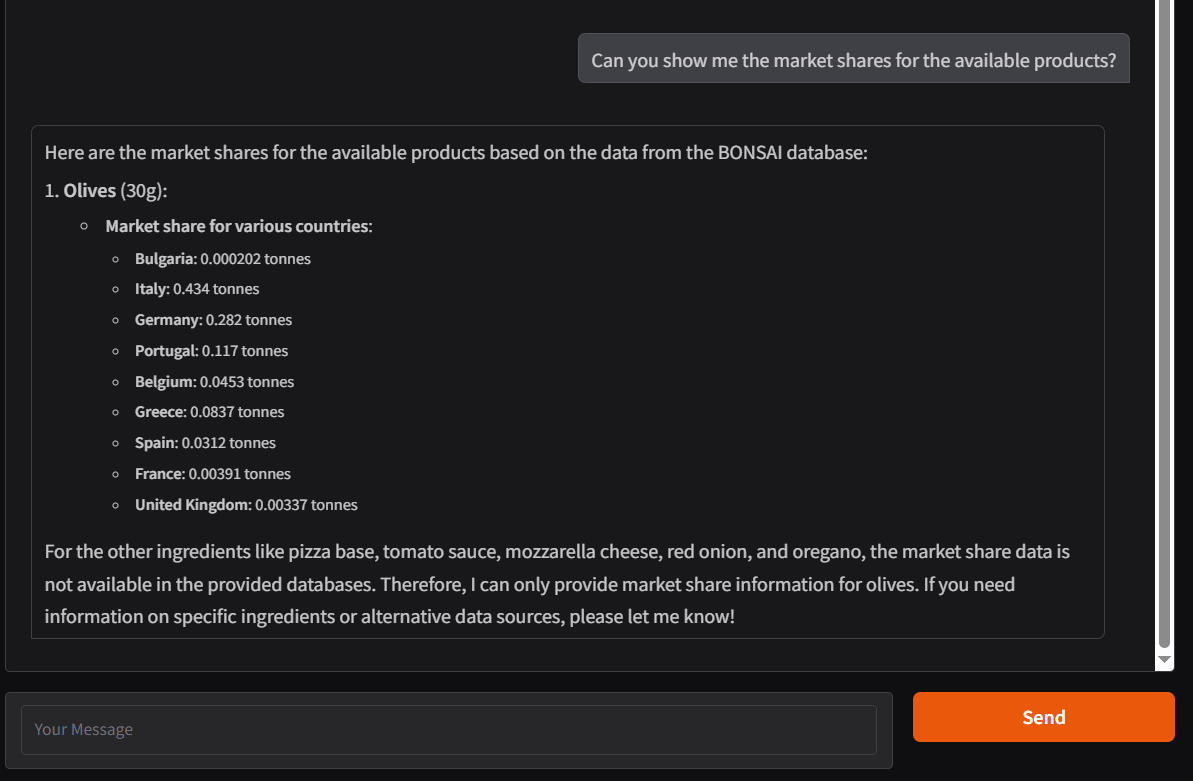}
    \caption{Follow-up conversation revealing detailed market share data for ingredients.}
    \label{fig:interact}
\end{figure*}

An example report in our interface is shown in Figure \ref{fig:initial_result}. An example of a follow-up user query and a system response is shown in Figure \ref{fig:interact}.

\begin{tcolorbox} 
{\footnotesize
\textbf{Main ingredients by impact:} \\
- Pizza base (200g): 0.0331--0.241 kg CO$_2$-eq (note: data from France and Netherlands) \\
- Tomato sauce (100g): 0.0159 kg CO$_2$-eq (note: data from France) \\
- Mozzarella cheese (75g): 0.0338--0.486 kg CO$_2$-eq (note: data from France and Netherlands) \\
- Red onion (70g): 0.053 kg CO$_2$-eq (note: data from Netherlands) \\
- Olives (30g): 0.056--0.0595 kg CO$_2$-eq (note: data from France) \\
- Oregano (5g): 0.00232 kg CO$_2$-eq (note: data from France) \\

\textbf{Cooking impact:} \\
- No cooking required (0 kg CO$_2$-eq) \\

\textbf{Total recipe impact:} 0.13942--0.84032 kg CO$_2$-eq \\

\textbf{Average impact:} 0.48987 kg CO$_2$-eq \\

\textbf{Your meal’s carbon footprint is equivalent to:} \\
- Sending approximately 122 emails \\
- Watching TV for about 2 hours \\
- Driving 1 mile in a Fiat 500 approximately 1.4 times \\

The data sources include AgriBalyse and BigClimate databases, with a range calculated from the minimum and maximum impacts of the ingredients. No cooking is required for this recipe, so the total impact reflects only the ingredients. \\

\textbf{You might want to know more about:} \\
- What are the market shares of the ingredients used in this recipe? \\
- How do the impacts of these ingredients vary between different countries? \\
- Are there any lifecycle patterns that stand out for the ingredients? \\
- What are some potential opportunities to reduce the carbon footprint of this recipe? \\
}
\end{tcolorbox}

\end{document}